\documentclass[10pt,twocolumn,letterpaper]{article}

\usepackage[pagenumbers]{cvpr} 

\definecolor{cvprblue}{rgb}{0.21,0.49,0.74}
\usepackage[pagebackref,breaklinks,colorlinks,citecolor=cvprblue]{hyperref}

\usepackage{multirow} 
\usepackage{colortbl}  
\usepackage{xcolor}
\usepackage{array}   
\usepackage{amsmath}
\usepackage{units} 
\usepackage{makecell}
\usepackage{arydshln}
\usepackage{booktabs}
\usepackage{graphicx}
\usepackage{nicematrix}
\usepackage{tabularx}
\usepackage{xfrac}
\usepackage{xspace}

\newcommand\Tstrut{\rule{0pt}{2.6ex}}         %
\newcommand\Bstrut{\rule[-0.9ex]{0pt}{0pt}}   %

\newcommand{\clip}{CLIP\xspace}

\newcommand{\oursfour}{FARE\textsuperscript{4}\xspace}
\newcommand{\ourstwo}{FARE\textsuperscript{2}\xspace}
\newcommand{\tecoafour}{TeCoA\textsuperscript{4}\xspace}
\newcommand{\tecoatwo}{TeCoA\textsuperscript{2}\xspace}

\definecolor{cvprblue}{rgb}{0.21,0.49,0.74}
\definecolor{lightgray}{rgb}{0.9,0.9,0.9}
\definecolor{BlueGray}{rgb}{1, 0.8, 0.8}
\definecolor{lightgreen}{rgb}{0.90, 0.99, 0.85}
\definecolor{darkgreen}{rgb}{.1, .85, .1}
\definecolor{newgray}{rgb}{0., 0., 0.} %
\definecolor{graygreen}{rgb}{.1, .75, .1} %
\definecolor{grayred}{rgb}{1., 0., 0.} %
\definecolor{lightorange}{rgb}{1., .9, 0.}
\definecolor{lighttred}{rgb}{1., .9, 0.8}
\definecolor{teaser1}{HTML}{FFCCBC}
\definecolor{teaser1}{HTML}{FFCCBC}

\newcolumntype{C}[1]{>{\centering\arraybackslash}p{#1}}
\newcolumntype{L}[1]{>{\raggedright\arraybackslash}p{#1}}
\newcolumntype{R}[1]{>{\raggedleft\arraybackslash}p{#1}}

\def\paperID{1814} 
\def\confName{CVPR}
\def\confYear{2025}

\title{OCRT: Boosting Foundation Models \\ in the Open World with Object-Concept-Relation Triad}

\author{Luyao Tang$^{1,2,*}$, Yuxuan Yuan$^{1}$$^{*}$, Chaoqi Chen$^{3,\dag}$, Zeyu Zhang$^{4}$, Yue Huang$^{1,2,\dag}$ and Kun Zhang$^{5,6}$\\
$^1$ Key Laboratory of Multimedia Trusted Perception and Efficient Computing, \\Ministry of Education of China, Xiamen University
{$^2$ School of Informatics, Xiamen University}\\
{$^3$ Shenzhen University}\quad
{$^4$ The Australian National University}\quad
{$^5$ Carnegie Mellon University}\\
{$^6$ Mohamed bin Zayed University of Artificial Intelligence}\\
{\tt\small  \{lytang, yuanyuxuan0908\}@stu.xmu.edu.cn, cqchen1994@gmail.com, steve.zeyu.zhang@outlook.com}\\
{\tt\small yhuang2010@xmu.edu.cn, kunz1@cmu.edu}
}

\begin{document}

\maketitle
\begin{abstract}
Although foundation models (FMs) claim to be powerful, their generalization ability significantly decreases when faced with distribution shifts, weak supervision, or malicious attacks in the open world. 
On the other hand, most domain generalization or adversarial fine-tuning methods are task-related or model-specific, ignoring the universality in practical applications and the transferability between FMs.
This paper delves into the problem of generalizing FMs to the out-of-domain data.
We propose a novel framework, the Object-Concept-Relation Triad (\textbf{OCRT}), that enables FMs to extract sparse, high-level concepts and intricate relational structures from raw visual inputs.
The key idea is to bind objects in visual scenes and a set of object-centric representations through unsupervised decoupling and iterative refinement. 
To be specific, we project the object-centric representations onto a semantic concept space that the model can readily interpret and estimate their importance to filter out irrelevant elements.
Then, a concept-based graph, which has a flexible degree, is constructed to incorporate the set of concepts and their corresponding importance, enabling the extraction of high-order factors from informative concepts and facilitating relational reasoning among these concepts.
Extensive experiments demonstrate that OCRT can substantially boost the generalizability and robustness of SAM and CLIP across multiple downstream tasks. \href{https://github.com/lytang63/OCRT}{Code}
\end{abstract}
\footnote{$^{*}$ Equal Contribution; $^{\dag}$ Corresponding Author.}
\section{Introduction}
\label{sec:intro}

\begin{figure}[tbp]
  \centering
  \subfloat[Segmentation with SAM]
  {\includegraphics[width=0.24\textwidth]{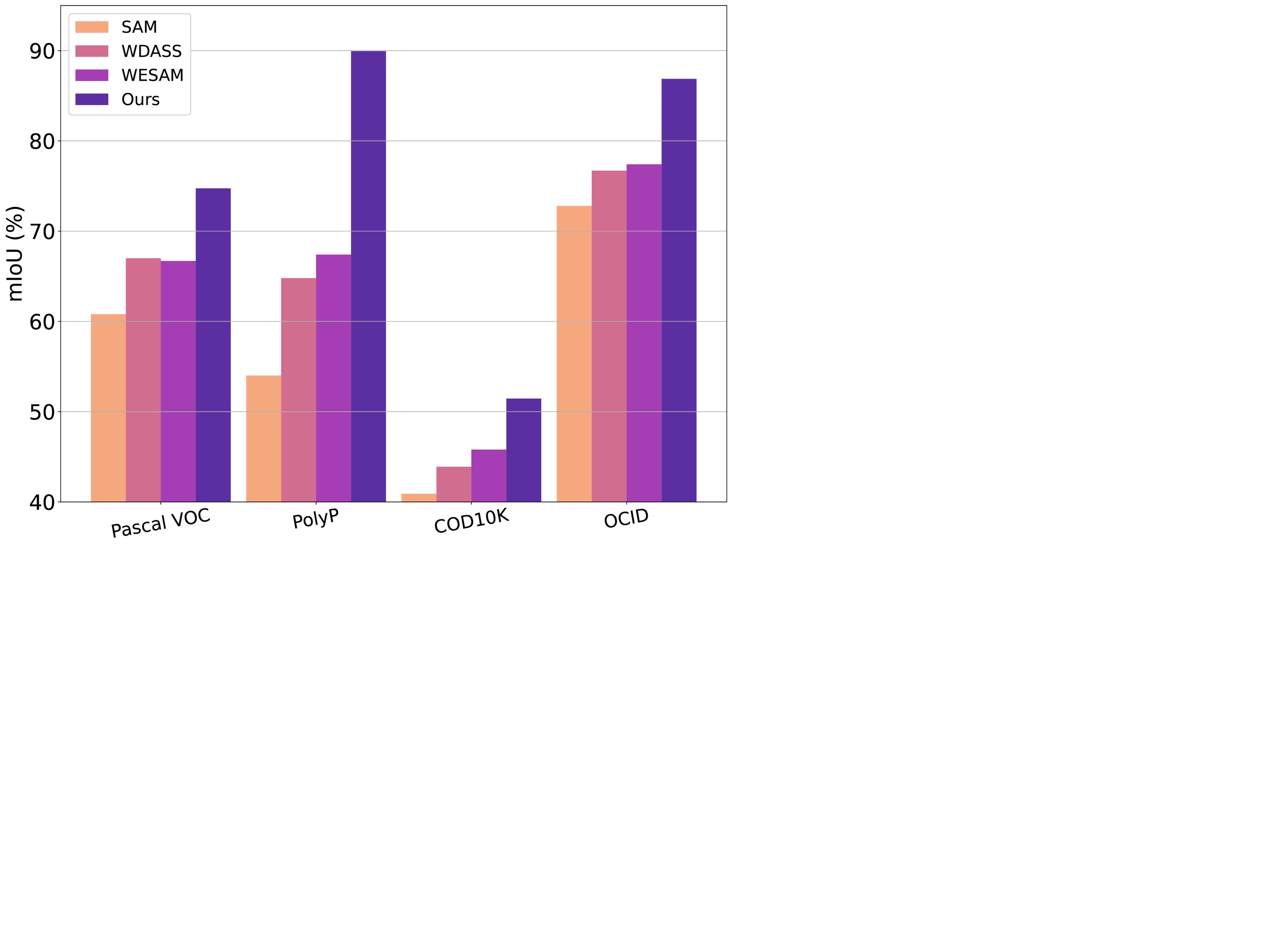}}
  \subfloat[Hallucination with CLIP]
  {\includegraphics[width=0.24\textwidth]{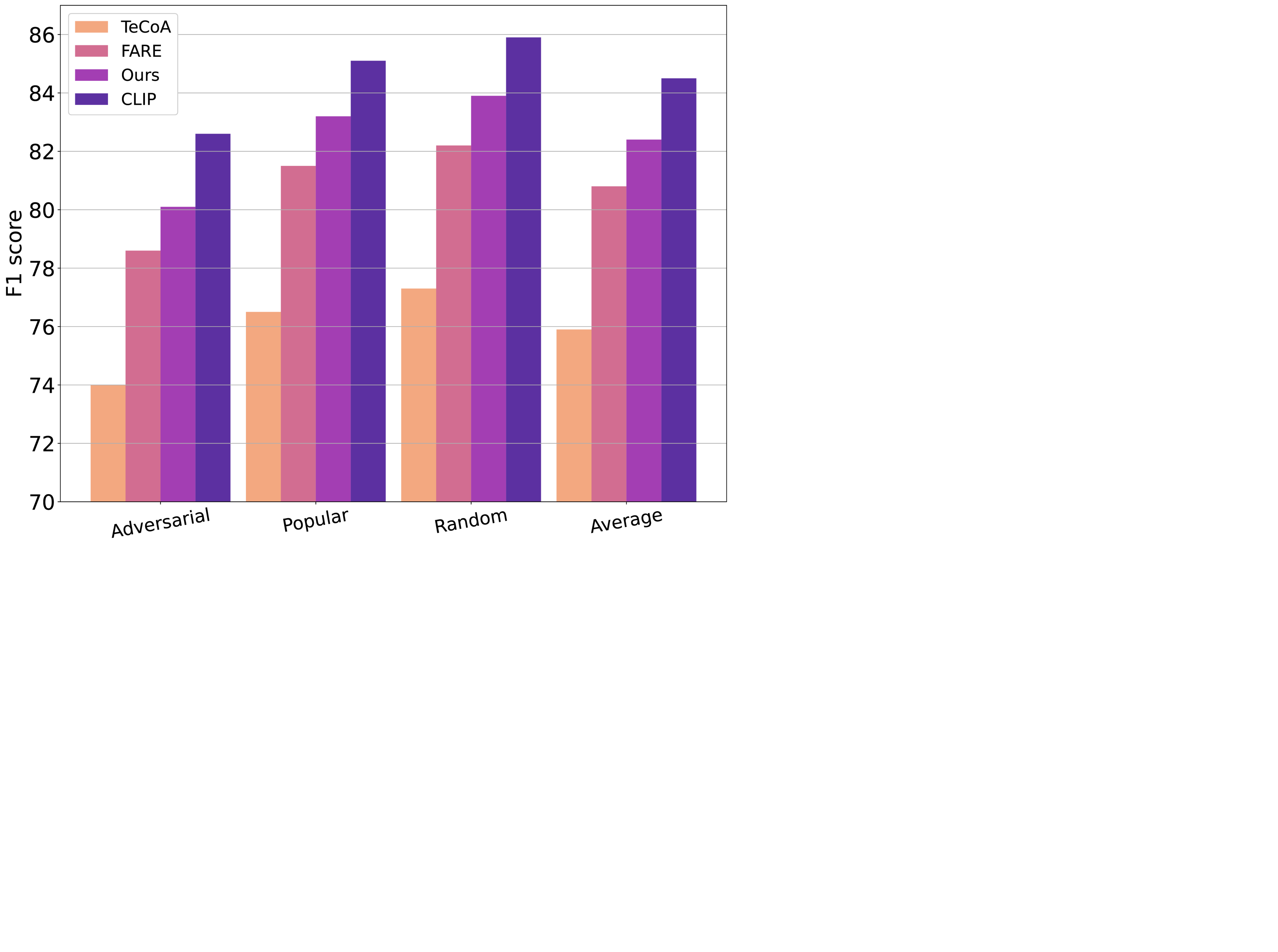}}
  \caption{Performance between SoTAs and OCRT across downstream tasks in the open world. (a) OCRT substantially resists performance damage from weak supervision. (b) OCRT mitigates the hallucination and comes closest to the original  CLIP.}
  \vspace{-5mm}
  \label{fig:lidar}
\end{figure}

The ability to generalize to unseen scenarios is considered an important sign of human intelligence~\cite{kejriwal2024challenges}. Foundation models (FMs)~\cite{achiam2023gpt,kirillov2023segment,radford2021learning} are trained on a large amount of single and multimodal data to bridge the gap between deep models and human intelligence. 
Although FMs claim to have achieved success in various downstream tasks, they are still highly vulnerable to out-of-domain data (\textit{distribution shift}~\cite{bommasani2021opportunities, zhang2021adaptive}, \textit{weak supervision}~\cite{yu2020fine, radford2023robust}, or \textit{malicious attacks}~\cite{zou2023universal}) in the open world, and this situation often occurs in real applications. For example, the vision foundation model (VFM)--SAM~\cite{kirillov2023segment} cannot accurately segment medical or camouflaged objects~\cite{chen2023sam, chen2025sam}. The multimodal foundation model (MMFM)--CLIP~\cite{radford2021learning} is extremely prone to attacks that can induce a wrong understanding of images~\cite{bai2024badclip, bansal2023cleanclip}. 
This inspires a series of works~\cite{zhou2024distilling, zhang2024improving,gao2024clip} to enhance the generalizability and robustness of FMs. 
They typically identify the generalized factors from a small amount of source domain data to fine-tune FMs and generalize the model to the out-of-domain data.

From the perspective of representation learning~\cite{van2017neural}, the mainstream paradigms for improving the generalizability of VFM and MMFM include inserting LoRA~\cite{hu2021lora} module or Adapter~\cite{gao2024clip,zhang2024segment} to fine-tune a small number of parameters, adversarial learning~\cite{li2024asam}, feature decomposition~\cite{ye2023large,drozdova2024semi}, and invariance alignment~\cite{singh2022flava}. Despite the general efficacy across different tasks, most existing approaches are customized for specific architectures and tasks. 
For instance, mask alignment~\cite{zhang2024improving, li2024asam} for SAM and semantic consistency~\cite{li2023clip, fan2024improving, wang2024diffusion} for CLIP are difficult to popularize, limiting the deployment of FMs in the wild~\cite{kejriwal2024challenges}.

To make FMs resilient to the out-of-domain data, we draw inspiration from human cognitive science~\cite{chen2005topological1,chen1982topological}. In particular, \cite{alexander2016relational,krawczyk2011hierarchy} points out that the core of human intelligence having high generalizability lies in that humans can extract intricate relational structures from dense sensory input~\cite{alexander2016relational}. For example, after the human eye receives visual signals in unfamiliar situations, the brain can decompose the scenes into regions of each object, and abstract their relationships to make logical reasoning and decisions accordingly~\cite{battaglia2018relational,krawczyk2012cognition}. 
Current FMs are sensitive to out-of-domain data for two reasons. First, due to low-level information decoupling, FMs cannot decompose visual scenes into unstructured entities (\textit{e.g.}, object regions) and fail to consider object structure. Second, regarding intricate relation extraction, FMs lack explicit processes for modeling relationships among these entities. \cite{ju2024survey} reveals FMs are vulnerable to low-order factors, leading to impaired generalizability.

Grounded on these insights, 
we propose a novel framework, the Object-Concept-Relation Triad (OCRT), to enhance the generalizability of FMs to out-of-domain data, particularly VFM and MMFM. 
(1) \textit{Object.} We first utilize a small amount of data from downstream tasks to bind objects in dense visual scenes and a set of object-centric representations, which divide the raw image into unstructured, low-level, and object-related regions. 
(2) \textit{Concept.} We project the object-centric representations onto a sparse and semantic concept space that the model can readily interpret, estimating their importance to filter out irrelevant entities.
(3) \textit{Relation.} We construct a concept-based graph that incorporates estimated concept importance and has a flexible degree, enabling the extraction of sparse and high-level factors from informative concepts and facilitating relational reasoning among these concepts.
The graph will be mapped to relation tokens and participate in fine-tuning FMs. 
OCRT does not incur any change to the FMs' internal architectures and works in a plug-and-play manner.

The main contributions are summarized as follows:

\begin{itemize}
\item We propose OCRT, a novel framework to enhance the generalizability and robustness of FMs\footnote{In this work, FMs are limited to VFM and MMFM.} caused by distribution shifts, weak supervision, and malicious attacks.

\item We transform the dense visual inputs into object-centric representations, which are subsequently projected into high-level concepts to facilitate relational reasoning. 

\item Extensive experiments on multiple downstream tasks demonstrate the superiority and scalability of OCRT when applied to SAM and CLIP models.

\end{itemize}

\section{Background and Motivation}
\label{sec:pre}

\begin{figure*}[t]
  \centering
   \includegraphics[width=0.95\linewidth]{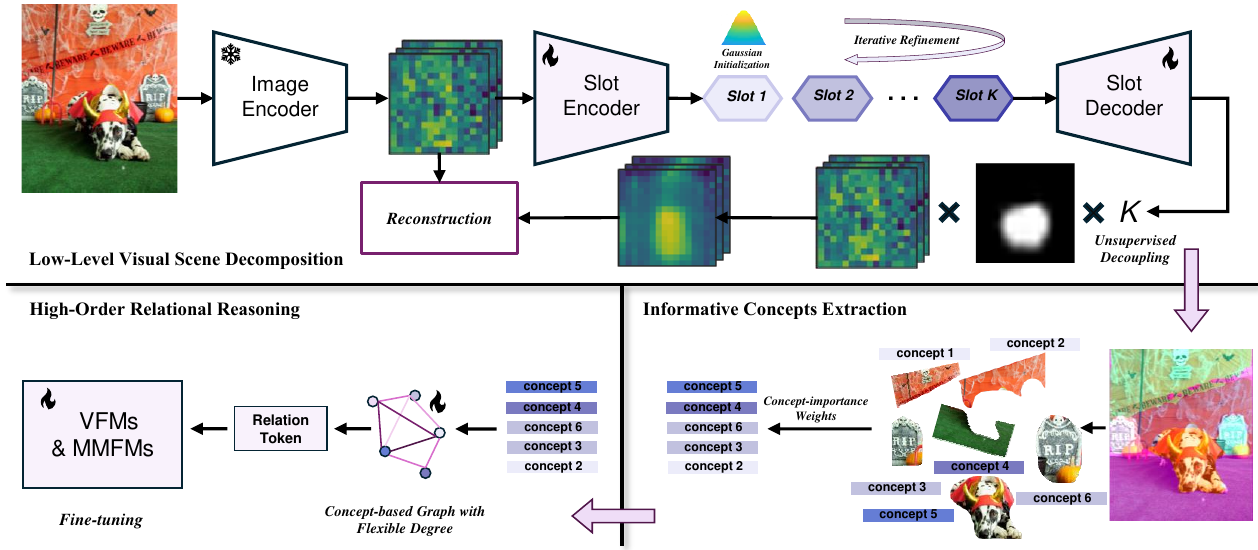}
   \caption{Overview of the proposed OCRT, which consists of three novel components: (1) Low-level visual scene decomposition with unstructured object-centric representations; (2) High-level informative concepts extraction via irrelevant concepts suppression; (3) Concept-based graph with flexible degree performs high-order relational reasoning for generalized factors.}
   \label{fig:overview}
\end{figure*}

\subsection{Problem Statement}
\label{subsec:notation}

We do not distinguish the optimization objectives of original or fine-tuned FMs and uniformly model them as $\mathcal{L}^{\text{base}}$.

\noindent
\textbf{SAM} as the representative of VFM. The main challenges faced by SAM in the open world are the \textit{distribution shift} and the \textit{weak supervision}. SAM has three main components: image encoder $\mathbf{z}=f_{\text{SAM}}(\mathbf{x} ; \Theta)$, prompt encoder $\mathbf{e}=g_{\text{SAM}}(\mathbf{p} ; \Omega)$, and mask decoder $\mathcal{M}=h(\mathbf{z}, \mathbf{e} ; \Phi)$. Inherited from the fine-tuned method~\cite{zhang2024improving}, we maintain anchor, student, and teacher encoders. The encoder outputs of strongly and weakly augmented views are $\mathbf{z}_s$ and $\mathbf{z}_w$, respectively. A base teacher-student self-training loss is:

\begin{equation}
\begin{aligned}
\mathcal{L}_{\text{SAM}}^{\text{base}}&=\mathcal{L}^{\text {dice}}\left(h(\mathbf{z}_{w/s}, \mathbf{e} ; \Phi_{stu/tea}), h(\mathbf{z}_{w}, \mathbf{e} ; \Phi_{anc})\right) \\
&+ \mathcal{L}^{\text {focal}}\left(h(\mathbf{z}_{w}, \mathbf{e} ; \Phi_{stu}), h(\mathbf{z}_{s}, \mathbf{e} ; \Phi_{tea})\right)  .
\label{eq:basesam}
\end{aligned}
\end{equation}

\noindent
\textbf{CLIP} as the representative of MMFM, with two main components, image encoder $\mathbf{z}=f_{\text{CLIP}}(\mathbf{x}; \Theta)$ and text encoder $\mathbf{e}=g_{\text{CLIP}}(\mathbf{e}; \Omega)$. The main challenges faced by CLIP in the open world are \textit{hallucination}, \textit{decreased robustness}, and \textit{impaired zero-shot performance} caused by malicious attacks. We conduct unsupervised adversarial fine-tuning~\cite{pmlr-v235-schlarmann24a}:

\begin{equation}
L_{\text{CLIP}}^{\text{base}}=\max _{\|\mathbf{x}^{_{\dag}}-\mathbf{x}\|_{\infty} \leq \varepsilon}\left\|f_{\text{CLIP}}(\mathbf{x}^{_{\dag}} ; \Theta^{_{\dag}})-\mathbf{z}\right\|_2^2 ,
\label{eq:baseclip}
\end{equation}

where $\mathbf{x}^{_{\dag}}$ represents perturbed data points. The loss $L_{\text{CLIP}}^{\text{base}}$ keeps $\mathbf{x}^{_{\dag}}$ close to the unperturbed $\mathbf{x}$ and $\left\|\mathbf{z}^{_{\dag}}-\mathbf{z}\right\|_2^2 \rightarrow 0$, the fine-tuned CLIP can be inserted into large VLMs, like LLaVA~\cite{liu2023llava, liu2024improved} to resist malicious attacks.

\subsection{Motivation of Object-Concept-Relation Triad}
\label{subsec:ocr}


The raw visual input $\mathbf{x}$ is dense, and it is decoupled into a finite set of low-level object-related factors $o^1, \cdots, o^K$, $o^i \in \mathcal{O}$~\cite{ferrari2007learning}. The generation process~\cite{wiles2021fine} of $\mathbf{x}$ from the deep models perspective is represented as

\begin{equation}
\begin{aligned}
&\mathbf{z} \sim p(\mathbf{z}), \quad o^i \sim p(o^i|\mathbf{z}), \quad \mathbf{x} \sim p(\mathbf{x}|\mathbf{z}),\\
&p(o^{1:K},\mathbf{x}) = p(o^{1:K})\int p(\mathbf{x}|\mathbf{z})p(\mathbf{z}|o^{1:K})d\mathbf{z}.
\end{aligned}
  \label{eq:object}
\end{equation}

The marginal distributions of $o^{1:K}$ in downstream tasks are different, but they share the same conditional generation process, resulting in a distribution shift  $p_{\text{train}}(o^{1:K}) \neq p_{\text{test}}(o^{1:K})$. Since the generation process is shared between the train and test distributions, we need to find object-related factors set $\tilde{\mathcal{O}} \subseteq \mathcal{O}$, that $p_{\text{train}}(\tilde{o}^{1:\tilde{K}},\mathbf{x}) = p_{\text{test}}(\tilde{o}^{1:\tilde{K}})\int p(\mathbf{x}|\mathbf{z}) 
 p(\mathbf{z}|\tilde{o}^{1:\tilde{K}})d\mathbf{z}$, to generalize models to the test distribution. The $\mathcal{O} \setminus \tilde{\mathcal{O}}$ indicates that not all object-related factors contribute to the generalization of the models. Specifically, we need to abstract $o^{1:K}$, extracting high-level and distribution-independent factors from $\mathcal{O}$. We consider them as the concepts set $c^{1:\tilde{K}}$, $c^i \in \mathcal{C}$ with

\begin{equation}
\begin{aligned}
p_{\text{train}}(c^{1:\tilde{K}},\mathbf{x}) = p_{\text{test}}(c^{1:\tilde{K}})\int p(\mathbf{x}|\mathbf{z}) 
 p(\mathbf{z}|c^{1:\tilde{K}})d\mathbf{z} .
\end{aligned}
  \label{eq:concept}
\end{equation}

Furthermore, the concepts only represent multiple single perspectives of $\mathbf{x}$, and they are discrete from each other. After the extraction of sparse and high-level concepts, the intricate relational reasoning~\cite{tian2024image} between concepts helps the model to generalize to the open world in line with human cognitive science~\cite{chen2005topological1,chen1982topological}. The formal representation is \underline{re}lational \underline{re}asoning with $\operatorname{ReRe}(c^{1:\tilde{K}}) \longrightarrow c^{1:\tilde{K}}$.

\section{Methodology}
\label{sec:method}

The proposed OCRT learning framework is illustrated in~\cref{fig:overview}. We aim to address the following questions: (1) how to decouple dense and low-level visual scenes into unstructured object-centric representations (\cref{subsec:objectcentric}), (2) how to estimate the importance of high-level informative concepts and extract the high-order relation from them (\cref{subsec:PCE} and \cref{subsec:ReRe}), and (3) how to use OCRT in real-world applications (\cref{subsec:realworld_app})?

\subsection{Low-Level Visual Scene Decomposition}
\label{subsec:objectcentric}

\noindent
\textbf{Iterative Refinement.} 
Inspired by how humans iteratively observe and think to discover an object's spatial location and semantics~\cite{chen2005topological1,chen1982topological}, we believe the refinement of objects' regions is iterative. With the Slot-Attention~\cite{locatello2020object}, we predefine a set of learnable object-centric representations $\mathbf{o}  \sim \mathcal{N}(\mathbf{o} ; \boldsymbol{\mu}, \boldsymbol{\sigma}) \in \mathbb{R}^{K \times D_o}$, where $K$ is the number of objects and $D_o$ is the number of the embedding's dimension.

To ensure semantic and spatial activation in FM's encoder output $\mathbf{z} \in \mathbb{R}^{N \times D_z}$ binds with object-centric representations $\mathbf{o}$, the attention mechanism~\cite{vaswani2017attention}, like human observation and thinking~\cite{hahn2020theoretical}, is implemented. For brevity, the outputs and projection networks (query, key, value) are represented as $\mathbf{k}=\mathcal{K}_\beta(\mathbf{z}) \in \mathbb{R}^{N \times D_o}$, $\mathbf{v}=\mathcal{V}_\phi(\mathbf{z}) \in \mathbb{R}^{N \times D_o}$, and $\mathbf{q}=\mathcal{Q}_\gamma(\mathbf{o}) \in \mathbb{R}^{K \times D_o}$, respectively. The \underline{refine}ment and \underline{att}e\underline{n}tion function are defined as:

\begin{equation}
\begin{aligned}
\operatorname{refine}(\boldsymbol{A}, \mathbf{v})&=\boldsymbol{A}^T \mathbf{v}, \quad A_{i j}=\frac{\operatorname{attn}(\mathbf{q}, \mathbf{k})_{i j}}{\sum_{l=1}^K \operatorname{attn}(\mathbf{q}, \mathbf{k})_{l j}}, \\
\operatorname{attn}(\mathbf{q}, \mathbf{k})&=\frac{e^{M_{i j}}}{\sum_{l=1}^N e^{M_{i l}}}, \quad \boldsymbol{M}=\frac{\mathbf{k q}^T}{\sqrt{D_o}},
\end{aligned}
  \label{eq:attn}
\end{equation}
where $\boldsymbol{A}$ is the attention matrix between $\mathbf{o}$ and $\mathbf{z}$. The step-by-step refinement is implemented by a Gated Recurrent Unit (GRU)~\cite{DBLP:journals/corr/ChoMGBSB14}. At the $t$-th iteration, the GRU accepts $\mathbf{o}^{t - 1}$ as input and the output is $\mathbf{o}^{t}=\operatorname{refine}\left(\operatorname{attn}\left(\hat{\mathbf{q}}^{t - 1}, \mathbf{k}\right), \mathbf{v}\right)$. The $\mathbf{o}^{0}$ is initialized with a Gaussian distribution~\cite{goodman1963statistical} to prevent iterative refinement with preference for certain objects.

\noindent
\textbf{Unsupervised Decoupling.} 
To bind $\mathbf{o}$ and the spatial region of each object in feature embedding $\mathbf{Z}$, we use a lightweight MLP as a spatial broadcast decoder $\operatorname{slot-decoder(\cdot)}$. $\operatorname{slot-decoder}$ decodes $k$-th $\mathbf{o}_k$ into an embedding with dimension of $\mathbb{R}^{N \times (D_z + 1)}$, which includes the activation region mask ${\alpha}_k \in \mathbb{R}^{N \times 1}$ and the reconstructed feature embedding $\mathbf{\hat{z}}_k \in \mathbb{R}^{N \times D_z}$ (semantic information in $\mathbf{z}$ helps the separation of $\mathbf{o}_k$ than reconstructing RGB pixels~\cite{seitzer2023bridging}). $\operatorname{softmax}$ is used to perform spatial dimension probability allocation on $\mathbf{o}$ of $K$ objects, and a competition between ${\alpha}_{1:K}$ is used to achieve unsupervised decoupling:

\begin{equation}
\mathbf{\hat{z}}=\sum_{k=1}^K \mathbf{\hat{z}}_k \odot \boldsymbol{m}_k, \quad \boldsymbol{m}_k=\underset{k}{\operatorname{softmax}}({\alpha}_k) .
\end{equation}

Minimizing reconstruction loss is used to align $\mathbf{\hat{z}}$ and $\mathbf{z}$:

\begin{equation}
\mathcal{L}_{\text {REC}}=\|\mathbf{\hat{z}}-\mathbf{z}\|^2, \quad \mathbf{\hat{z}}=\operatorname{slot-decoder}(\mathbf{o}) .
\end{equation}

\subsection{Informative Concepts Extraction} 
\label{subsec:PCE}
The $\mathbf{o}$ obtained in~\cref{subsec:objectcentric} can bind objects in visual scenes and a set of object-centric representations. But for downstream tasks like segmentation and semantic understanding, the low-level object-centric representations cannot be interpreted~\cite {koh2020concept} by FMs. We need to map them to the semantic concept space $\mathcal{C}$ related to the task. First, for concept generation, we know from existing research~\cite{zarlenga2022concept} that a set of explanatory and semantically rich embeddings can be regarded as concepts. From the perspective that the FMs can readily interpret, concepts set $c^1, \cdots, c^{K}$ extracted from object-centric representations are $\mathbf{c}_{1:K} = \mathbf{z}_{1:K} \odot \boldsymbol{m}_{1:K}$.


Due to low-level factors such as background, occlusion, and illumination changes, not all $K$ concepts are task-related and informative. We propose Informative Concepts Extraction (Irrelevant Concepts Suppression) to extract task-related concepts, which widely exist in embeddings~\cite{zarlenga2022concept}. The concept-importance weight of the $k$-th concept is estimated from $\mathbf{c}\in \mathbb{R}^{N \times K \times D_z}$

\begin{equation}
\omega_{k}=\frac{1}{K}\sum_{j = 1}^{K}\frac{\boldsymbol{c}_{k}\cdot\boldsymbol{c}_{j}}{\lVert\boldsymbol{c}_{k}\rVert_{2}\lVert\boldsymbol{c}_{j}\rVert_{2}} .
\label{eq:omega}
\end{equation}

\begin{figure}[t]
  \centering
   \includegraphics[width=0.95\linewidth]{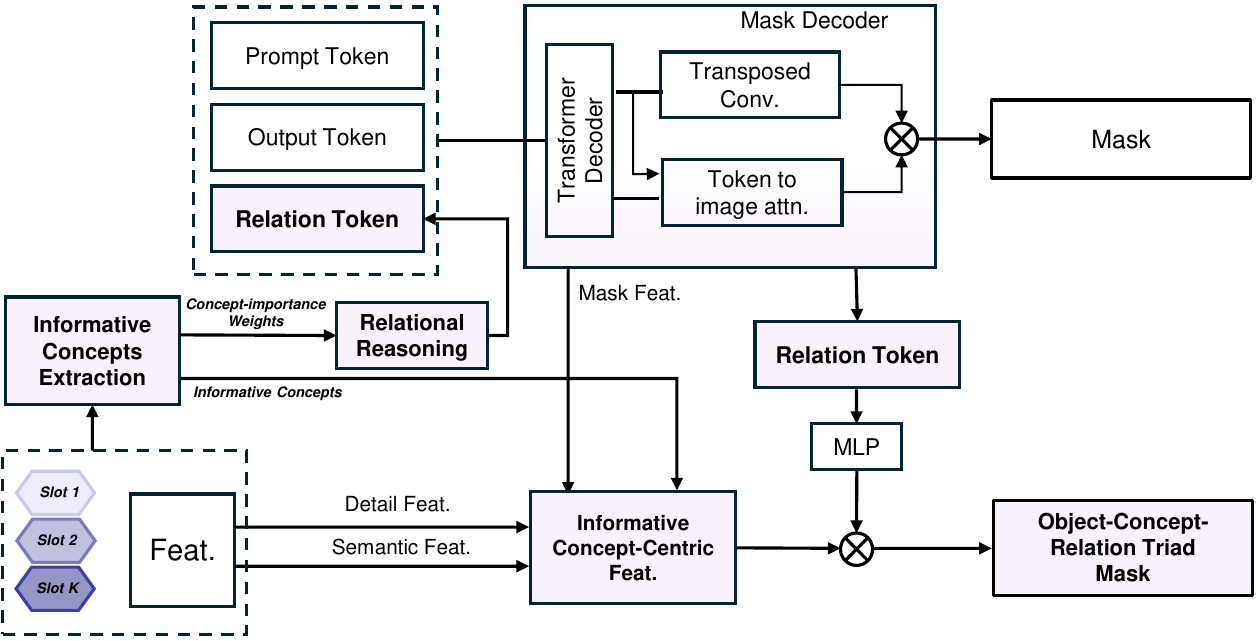}
   \caption{Object-Concept-Relation Triad Decoder of SAM.}
   \label{fig:decoder}
\end{figure}

For VFM, we extract the detailed feature after the first attention block of the encoder as shown in~\cref{fig:decoder}. After transposed convolution, detail and semantic features are element-wise added to obtain object features $\mathbf{z}_{obj}$. Furthermore, we introduce importance weight $\omega_{k}$ in~\cref{eq:omega} for informative concepts extraction to map $\mathbf{z}_{obj}$ in low-level object space to features in semantic concept space:

\begin{equation}
\mathbf{z}_{cpt} = \mathbf{z}_{obj} \odot \sum_{i = 1}^{\tilde{K}}\mathbb{I}_{\left[k \in \operatorname{Top}^{\tilde{K}}(\omega_{1:K})\right]} \boldsymbol{m}_k ,
\label{eq:concept}
\end{equation}
where $\mathbb{I}_{\left[ \cdot \right]}$ is an indicator function and $\operatorname{Top}^{\tilde{K}}(\cdot)$ lists the informative concepts with the highest importance $\tilde{K}$. $K-\tilde{K}$ represents the number of suppressed irrelevant concepts.


\begin{table*}[t]
  \centering
       \scalebox{0.90}
{
    \begin{tabular}{c|c|c|c|c|c|c|c|c|c|c|c|c}
    \hline \multirow{2}{*}{ Method } & \multicolumn{3}{c|}{ \cellcolor{blue!10}COCO 2017 } & \multicolumn{3}{c}{ \cellcolor{blue!10}Pascal VOC } & \multicolumn{3}{|c|}{ \cellcolor{green!10}kvasir-SEG } & \multicolumn{3}{|c}{ \cellcolor{green!10}ISIC }\\
    & \cellcolor{gray!15}box & \cellcolor{gray!30}point & \cellcolor{gray!45}poly & \cellcolor{gray!15}box & \cellcolor{gray!30}point & \cellcolor{gray!45}poly & \cellcolor{gray!15}box & \cellcolor{gray!30}point & \cellcolor{gray!45}poly & \cellcolor{gray!15}box & \cellcolor{gray!30}point & \cellcolor{gray!45}poly\\
    \hline 
    SAM~\citep{kirillov2023segment} & 74.29 & 55.06 & 65.64 & 69.21 & 69.21 & 60.79 & 81.59 & 62.30 & 54.03 & 66.74 & 53.42 & 62.82\\
    \hline 
    TENT~\citep{wang2021tent} & 78.21 & 52.99 & 71.51 & 80.24 & 74.97 & 65.03  & 82.47 & 61.84 & 62.97 & 71.76 & 53.46 & 67.12\\
    SHOT~\citep{liang2021source} & 75.18 & 58.46 & 69.26 & 79.80 & 74.26 & 63.38 & 82.30 & 63.76 & 61.34 & 71.99 & 55.99 & 66.86\\
    TRIBE~\citep{su2024towards} & 77.56 & 49.56 & 70.99 & 78.87 & 69.21 & 65.39 & 85.05 & 73.03 & 64.61 & 72.61 & 50.36 & 67.99\\
    DePT~\citep{gao2022visual} & 71.00 & 37.35 & 63.27 & 74.09 & 42.99 & 59.94 & 81.91 & 52.06 & 61.55 & 78.43 & 46.79 & 72.75\\
    WDASS~\citep{das2023weakly} & 77.29 & 60.55 & 70.19 & 80.12 & 76.15 & 66.98 & 84.01 & 63.78 & 64.78 & 74.23 & 55.63 & 67.84\\
    WESAM$\dagger$~\citep{zhang2024improving} & 77.32 & 60.50 & 70.77 & 80.27 & 74.15 & 66.72 & 85.47 & 75.23 & 67.40 & 80.01 & 62.12 & 75.36\\
    \hline
    Ours & $\mathbf{78.74}$ & $\mathbf{63.82}$ & $\mathbf{75.60}$ & $\mathbf{83.63}$ & $\mathbf{78.91}$ & $\mathbf{74.74}$ & $\mathbf{89.95}$ & $\mathbf{84.69}$ & $\mathbf{89.94}$ & $\mathbf{83.82}$ & $\mathbf{66.89}$ & $\mathbf{78.83}$\\
    $\triangle$ over SAM & +4.45 & +8.76 & +9.96 & +14.42 & +9.70 & +13.95 & +8.36 & +22.39 & +35.91 & +17.08 & +13.47 & +16.01\\
    $\triangle$ over WESAM& +1.42 & +3.32 & +4.83 & +3.36 & +4.76 & +8.02 & +4.48 & +9.46 & +22.54 & +3.81 & +4.77 & +3.47\\
    \hline 
    Supervised & 81.50 & 69.77 & 73.39 & 81.23 & 76.98 & 71.32 & 85.89 & 77.54 & 81.64 & 81.62 & 79.81 & 80.26\\
    \hline
    \end{tabular}
}
  \caption{Comparison on \colorbox{blue!10}{\color{black}natural} and \colorbox{green!10}{\color{black}medical} images with \colorbox{gray!15}{\color{black}bounding box}, \colorbox{gray!30}{\color{black}sparse points}, and \colorbox{gray!45}{\color{black}coarse mask} prompts.\label{tab:natural_medical}}
\end{table*}

\subsection{High-Order Relational Reasoning}
\label{subsec:ReRe}


In the process of low-level visual scene decomposition and extraction of high-level informative concepts, $\mathbf{o}$ is directly bound to a certain region of the visual scene through $\boldsymbol{m}$, $\mathbf{o}_k$ and $\mathbf{o}_{k^{_{'}}}$ are rigidly separated at the spatial and semantic levels, making image modeling uneven. It ignores the native high-order topological correlation of semantic space, which is unfavorable for downstream tasks (such as segmentation and semantic understanding) that need to perceive the intricate relation in the entire image.

\begin{figure}[h]
  \centering
   \includegraphics[width=0.95\linewidth]{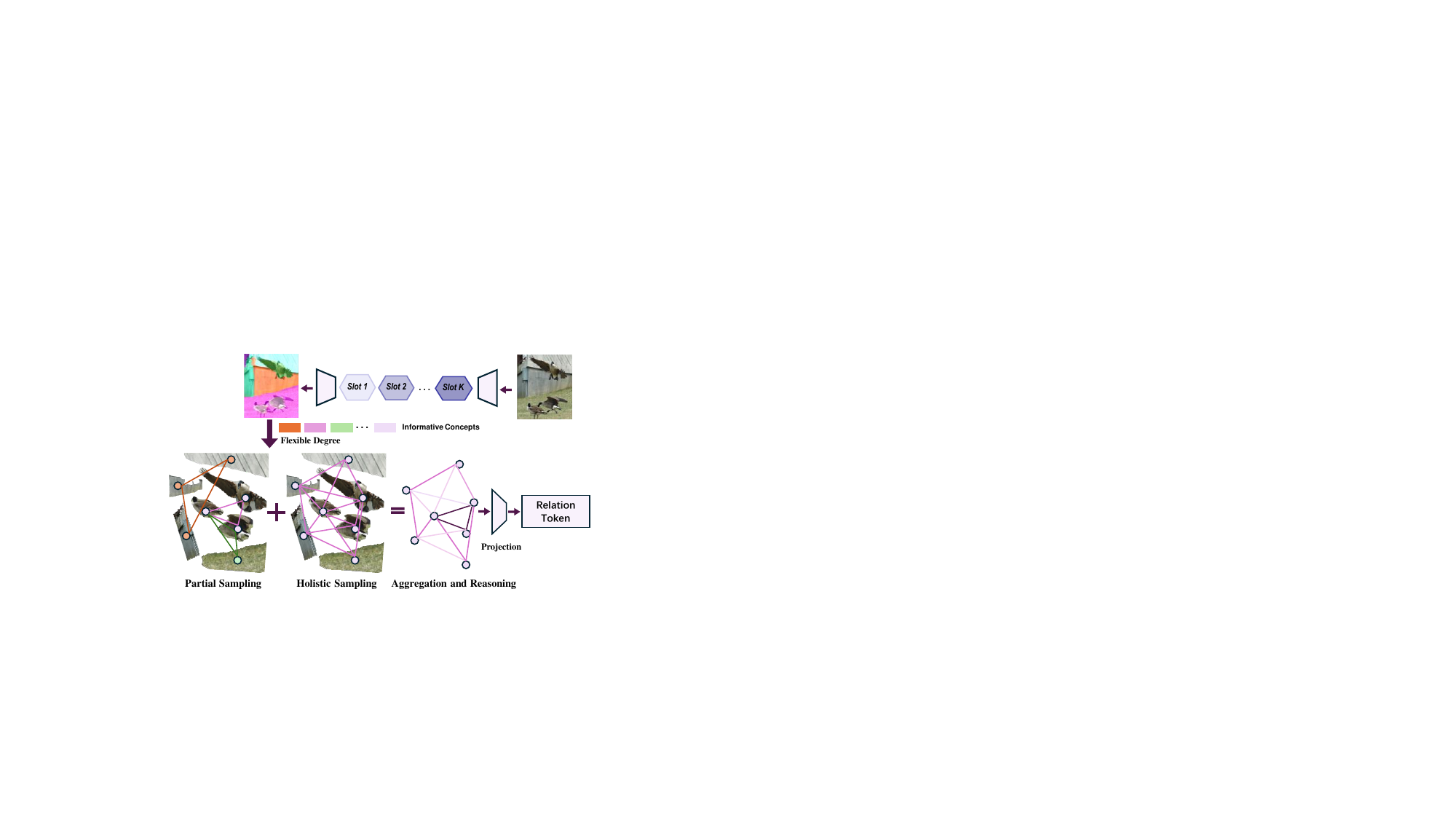}
   \caption{Concept-based graph with flexible degree.}
   \label{fig:graph}
\end{figure}

To break the rigidity, the challenging task is to extract high-order relationships from a dynamic and scalable perspective~\cite{li2023sigma++}. We propose a concept-based graph to break through the extraction process so that $\mathbf{o}_{k/k^{_{'}}}$ can communicate and perform gathering and distributing information flexibly. Eventually, the graph is aggregated into task-related Relation Tokens $\mathcal{T}_{rel} \in \mathbb{R}^{N_r \times D_r}$.

For \textbf{degree}, to overcome the problem of biased information transmission caused by the rigid limitation of degree by traditional graphs, the degree budgets~\cite{tian2024image} are assigned to nodes $o$\footnote{For brevity, we regard nodes $o$ as a set of embeddings $\mathbf{o}$.} based on $w$ in~\cref{eq:omega}, that $o$ is in proportion to its corresponding concept-importance: $\operatorname{deg}(o) \propto w_{1:K}$.

For \textbf{nodes}, similar to the human process~\cite{chen2005topological1,chen1982topological} of perceiving from local to overall and thinking divergently first and then reasoning, we pursue nodes' flexibility. As shown in~\cref{fig:graph}, regional and global information is aggregated by sampling partial and holistic nodes, respectively.

For \textbf{aggregation}, we perform the aggregation guided by concept-importance. The $i$-th layer in the concept-based graph (formally equal to $\operatorname{ReRe(\cdot)}$ in~\cref{subsec:ocr}) accepts embeddings $\mathbf{o}^{i-1}$ as input, and the set of neighbor nodes is $\mathcal{N}(o)$, then the relational output of node $o$ is

\begin{equation}
\mathbf{z}_o^i = \operatorname{ReRe}(\mathbf{z}_o^{i-1}) = \frac{1}{C^i} \sum_{o^{{_{'}}} \in \mathcal{N}(o)} \exp \left(\operatorname{sim}(o^{{_{'}}}, o)\right) \mathbf{z}_{o^{{_{'}}}}^{i-1} ,
\end{equation}
where $\operatorname{sim}(\cdot,\cdot)$ is a parameterized function, such as cosine similarity, used to measure the correlation between node pairs ($o$, $o^{{_{'}}}$). $C^i$ is a normalized constant. The $\mathbf{z}_o$ of the last concept-based graph layer is projected into Relation Tokens $\mathcal{T}_{rel} \in \mathbb{R}^{N_r \times D_r}$ as shown in~\cref{fig:graph,fig:decoder}.

\subsection{Real-Wolrd Applications}
\label{subsec:realworld_app}

\textbf{Boosting on SAM.} Self-training networks may have error accumulation due to incorrect predictions. We fix the parameters of the anchor model (input: $\mathbf{x}_w$, output: $\mathcal{M}^a$). The trained model is called the OCRT model (input: $\mathbf{x}_s$, output: $\mathcal{M}$). We use loss function in the style of $\mathcal{L}^{\text {base}}$~\cite{zhang2024improving} to train the Relational Reasoning and Fusion modules to prevent significant bias in knowledge transfer: $\mathcal{L}_{\text{SAM}}^{\text{OCRT}} =\mathcal{L}^{\text {bce}}\left(\mathcal{M}, \mathcal{M}^a\right) + \mathcal{L}^{\text {dice}}\left(\mathcal{M}, \mathcal{M}^a\right)$. In the fine-tuning stages, we use a bootstrap strategy. At the end of each epoch in which the model improves on the validation set, we directly copy the parameters of the OCRT model to the anchor model. Through this iterative process, we gradually complete the boosting of the VFM.

\noindent
\textbf{Boosting on CLIP.} 
Our purpose is to fine-tune CLIP's image encoder to make it robust against adversarial attacks in the open world with unaffected output and zero-shot performance for clean samples. Humans naturally have a robust extraction ability~\cite{krawczyk2012cognition} for noisy and occluded images because the brain's extraction of unstructured information can assist humans in judging the semantics of polluted images from high-order relations~\cite{zhou2018temporal, santoro2017simple}. CLIP needs to be able to perform high-order relational reasoning between unstructured information to resist attacks more effectively, like humans~\cite{krawczyk2012cognition}. This is in line with our scheme's main thrust. We seamlessly integrate the Relation Token $\mathcal{T}_{rel}$ in~\cref{subsec:objectcentric} into the unsupervised adversarial fine-tuning of CLIP~\cite{pmlr-v235-schlarmann24a}: $L_{\text{CLIP}}^{\text{OCRT}}=\max _{\|\mathbf{x}^{_{\dag}}-\mathbf{x}\|_{\infty} \leq \varepsilon}\left\|f_{\text{CLIP}}(\mathbf{x}^{_{\dag}}, \mathcal{T}_{rel} ; \Theta^{_{\dag}}_{img})-\mathbf{z}_{img}\right\|_2^2$, where we concatenate $\mathcal{T}_{rel}$ with the original tokens. By sharing parameters in attention and FFN between tokens, the global context and high-order information propagate~\cite{kreuzer2021rethinking} with each other, the MMFM captures more high-order factors to resist attacks in the open world.


\section{Experiments}
\label{sec:exp}

\begin{table*}[!t]
\centering
\small
\tabcolsep=2.7pt
\extrarowheight=-0.8pt
\begin{tabular}{C{5mm} C{25mm} | C{6.5mm} | C{6.5mm} C{6.5mm} *{3}{C{6.5mm} C{6.5mm} C{6.5mm}} C{6.5mm} C{6.5mm} | C{6.5mm} C{6.5mm}}
\bottomrule
Eval. & \makecell{Vision\\ Encoder} & \parbox[c][1.6cm]{0.8cm}{\centering{\rotatebox[origin=c]{90}{ImageNet}}} & \parbox[c][1.2cm]{0.8cm}{\centering{\rotatebox[origin=c]{90}{CalTech}}} & \parbox[c][1.2cm]{0.8cm}{\centering\rotatebox[origin=c]{90}{Cars}} & \parbox[c][1.2cm]{0.8cm}{\centering\rotatebox[origin=c]{90}{CIFAR10}} & \parbox[c][1.2cm]{0.8cm}{\centering\rotatebox[origin=c]{90}{CIFAR100}} & \parbox[c][1.2cm]{0.8cm}{\centering\rotatebox[origin=c]{90}{DTD}} & \parbox[c][1.2cm]{0.8cm}{\centering\rotatebox[origin=c]{90}{EuroSAT}} & \parbox[c][1.2cm]{0.8cm}{\centering\rotatebox[origin=c]{90}{FGVC}} & \parbox[c][1.2cm]{0.8cm}{\centering\rotatebox[origin=c]{90}{Flowers}} &  \parbox[c][1.2cm]{0.8cm}{\centering\rotatebox[origin=c]{90}{ImageNet-R}} & \parbox[c][1.2cm]{0.8cm}{\centering\rotatebox[origin=c]{90}{ImageNet-S}}& \parbox[c][1.2cm]{0.8cm}{\centering\rotatebox[origin=c]{90}{PCAM}} &\parbox[c][1.2cm]{0.8cm}{\centering\rotatebox[origin=c]{90}{OxfordPets}} & \parbox[c][1.2cm]{0.8cm}{\centering\rotatebox[origin=c]{90}{STL-10}} & 
\multicolumn{2}{c}{\makecell{Avg.}}
\\
\hline
\parbox[t]{3mm}{\multirow{4}{*}{\rotatebox[origin=c]{90}{clean}}}
& {\cellcolor{lightgray}}CLIP & {\cellcolor{lightgray}}74.9 &{\cellcolor{lightgray}} 83.3 & {\cellcolor{lightgray}}77.9 &{\cellcolor{lightgray}} 95.2 & {\cellcolor{lightgray}}71.1 & {\cellcolor{lightgray}}55.2 & {\cellcolor{lightgray}}62.6 & {\cellcolor{lightgray}}31.8 & {\cellcolor{lightgray}}79.2 & {\cellcolor{lightgray}}87.9 & {\cellcolor{lightgray}}59.6 &{\cellcolor{lightgray}} 52.0 & {\cellcolor{lightgray}}93.2 &{\cellcolor{lightgray}} 99.3 & {\cellcolor{lightgray}}73.1 &{\cellcolor{lightgray}}\\
& \tecoafour & 75.2 & 78.4 & 37.9 & 79.6 & 50.3 & 38.0 & 22.5 & 11.8 & 38.4 & 74.3 & 54.2 & 50.0 & 76.1 & 93.4  & 54.2 &\\ 
& \oursfour & 70.4 & 84.7 & 63.8 & 77.7 & 56.5 & 43.8 & 18.3 & 22.0 & 58.1 & 80.2 & 56.7 & 50.0 & 87.1 & 96.0  & 61.1 & \\
& OCRT\textsuperscript{4} & 73.6 & 85.9 & 65.2 & 78.0 & 57.1 & 43.9 & 19.3 & 23.1 & 60.0 & 83.5 & 58.4 & 50.9 & 88.8 & 96.4  & 63.2 & \scriptsize{\textcolor{blue}{$\uparrow$2.1}}\\
\hline

\parbox[t]{3mm}{\multirow{4}{*}{\rotatebox[origin=c]{90}{$\varepsilon=\nicefrac{2}{255}$}}}
& {\cellcolor{lightgray}}\clip & {\cellcolor{lightgray}}0.0 & {\cellcolor{lightgray}}0.0 &{\cellcolor{lightgray}}0.0 &{\cellcolor{lightgray}}0.0 &{\cellcolor{lightgray}}0.0 &{\cellcolor{lightgray}}0.0 &{\cellcolor{lightgray}}0.0 &{\cellcolor{lightgray}}0.0 &{\cellcolor{lightgray}}0.0 &{\cellcolor{lightgray}}0.0 &{\cellcolor{lightgray}}0.1 &{\cellcolor{lightgray}}0.0 &{\cellcolor{lightgray}}0.0 &{\cellcolor{lightgray}}0.0 &{\cellcolor{lightgray}}0.0 & {\cellcolor{lightgray}}\\
& \tecoatwo & 62.3 & 70.2 & 22.2 & 63.7 & 35.0 & 27.0 & 12.8 & 5.8 & 27.6 & 58.8 & 45.2 & 40.0 & 69.7 & 88.7 & 43.6 & \\
& \ourstwo & 46.1 & 73.0 & 26.0 & 60.3 & 35.6 & 26.7 & 6.2 & 5.9 & 31.2 & 56.5 & 38.3 & 41.9 & 68.3 & 90.1  & 43.1 &  \\
& OCRT\textsuperscript{2} & 48.7 & 75.9 & 27.1 & 61.0 & 37.5 & 28.9 & 9.2 & 9.8 & 33.9 & 59.8 & 38.6 & 43.8 & 69.4 & 91.8  & 45.4 &\scriptsize{\textcolor{blue}{$\uparrow$1.8}} \\
\hline
\parbox[t]{3mm}{\multirow{4}{*}{\rotatebox[origin=c]{90}{$\varepsilon=\nicefrac{4}{255}$}}}& {\cellcolor{lightgray}}\clip & {\cellcolor{lightgray}}0.0 &{\cellcolor{lightgray}}0.0&	{\cellcolor{lightgray}}0.0	&{\cellcolor{lightgray}}0.0	&{\cellcolor{lightgray}}0.0	&{\cellcolor{lightgray}}0.0	& {\cellcolor{lightgray}}0.0 &	{\cellcolor{lightgray}}0.0	&{\cellcolor{lightgray}}0.0		&{\cellcolor{lightgray}}0.0	&{\cellcolor{lightgray}}0.0	&{\cellcolor{lightgray}}0.0	&{\cellcolor{lightgray}}0.0	&{\cellcolor{lightgray}}0.0	& {\cellcolor{lightgray}}0.0&{\cellcolor{lightgray}} \\
& \tecoafour & 44.3 & 60.9 & 8.4 & 37.1 & 21.5 & 16.4 & 6.6 & 2.1 & 12.4 & 41.9 & 34.2 & 44.0 & 55.2 & 74.3  & 31.9 & \\
& \oursfour & 33.3 & 64.1 & 12.7 & 34.6 & 20.2 & 17.3 & 11.1 & 2.6 & 12.5 & 40.6 & 30.9 & 50.2 & 50.7 & 74.4  & 32.4 & \\
& OCRT\textsuperscript{4} & 35.6 & 65.7 & 12.9 & 35.9 & 21.0 & 17.7 & 12.0 & 2.7 & 12.8 & 41.2 & 31.8 & 51.4 & 51.7 & 74.6  & 33.4 & \scriptsize{\textcolor{blue}{$\uparrow$1.0}}\\
\bottomrule
\end{tabular} 
   \caption{Clean and adversarial evaluation on zero-shot image classification datasets of CLIP model.\label{tab:zero_shot}}
\end{table*}


Through the experimental evaluation, we aim to answer the following questions. (1) Can OCRT generalize consistently to small and significant distribution shifts? (2) How is the transferability of OCRT and can it be versatile between VFM and MMFM, and out-of-domain tasks? (3) How sensitive is OCRT to hyperparameters and what are the data assumptions for tuning it?

\subsection{Experiments on SAM}
\label{subsec:exp_setup}

\noindent
\textbf{Benchmarks.} 
\textit{Datasets}: The source domain of SAM is natural images dataset SA-1B~\cite{kirillov2023segment}. Following previous works~\cite{chen2023sam, zhang2024improving}, target domains are: natural images with small distribution shift (VOC~\cite{Everingham15}, COCO~\cite{lin2014microsoft}), robotic, camouflaged and medical images with significant distribution shift (OCID~\cite{DBLP:conf/icra/SuchiPFV19}, CAMO~\cite{le2019anabranch}, COD10K~\cite{fan2020camouflaged} and ISIC~\cite{codella2018skin}, kvasir-SEG~\cite{jha2020kvasir}). \textit{Weak supervision}: bounding \textbf{box}, \textbf{point} with $5$ in/out of instance and \textbf{poly} with $P/20$ vertices, where $P$ is the mask's perimeter.


\noindent
\textbf{Baselines.} 
TENT~\cite{wang2021tent}, TRIBE~\cite{su2024towards}, SHOT~\cite{liang2021source} adapt the model to continuous domain shifts. WDASS~\cite{das2023weakly}, WESAM~\cite{zhang2024improving}, and DePT~\cite{gao2022visual} are for the adaptation of weakly supervised domains. Supervised fine-tuning SAM with ground-truth masks. $\dagger$ denotes reproduced results.


\begin{table}[t]
  \centering
       \scalebox{0.55}
{
    \begin{tabular}{c|c|c|c|c|c|c|c|c|c}
    \hline \multirow{2}{*}{ Method } & \multicolumn{3}{c}{ \cellcolor{yellow!10}CAMO } & \multicolumn{3}{|c|}{ \cellcolor{yellow!10}COD10K } & \multicolumn{3}{c}{ \cellcolor{red!10}OCID }\\
    & \cellcolor{gray!15}box & \cellcolor{gray!30}point & \cellcolor{gray!45}poly & \cellcolor{gray!15}box & \cellcolor{gray!30}point & \cellcolor{gray!45}poly & \cellcolor{gray!15}box & \cellcolor{gray!30}point & \cellcolor{gray!45}poly\\
    \hline 
    SAM~\citep{kirillov2023segment} & 62.72 & 57.43 & 50.85 & 66.32 & 63.61 & 40.04 & 86.35 & 71.41 & 72.81\\
    \hline 
    TENT~\citep{wang2021tent} & 71.24 & 59.59 & 60.29 & 69.36 & 61.94 & 43.36 & 87.77 & 66.61 & 77.53\\
    SHOT~\citep{liang2021source} & 71.61 & 62.78 & 58.72 & 69.09 & 65.25 & 42.38 & 88.06 & 74.39 & 76.25  \\
    TRIBE~\citep{su2024towards} & 66.00 & 61.97 & 60.54 & 67.84 & 63.62 & 42.75 & 86.77 & 67.86 & 76.50 \\
    DePT~\citep{gao2022visual} & 55.44 & 33.07 & 48.63 & 59.32 & 34.06 & 35.51 & 82.00 & 56.52 & 70.92\\
    WDASS~\citep{das2023weakly} & 71.25 & 63.39 & 62.29 & 71.42 & 65.61 & 43.93 & 87.68 & 77.13 & 76.70\\
    WESAM$\dagger$~\citep{zhang2024improving} & 73.42 & 65.55 & 62.90 & 71.93 & 70.55 & 45.87 & 88.09 & 80.14 & 77.41 \\
    \hline
    Ours & $\mathbf{76.32}$ & $\mathbf{73.81}$ & $\mathbf{71.03}$ & $\mathbf{74.41}$ & $\mathbf{70.99}$ & $\mathbf{51.45}$ & $\mathbf{88.43}$ & $\mathbf{80.95}$ & $\mathbf{86.87}$ \\
     $\triangle$ over SAM& +13.60 & +16.38 & +20.18 & +8.09 & +6.38 & +11.50 & +2.08& +8.54 & +14.06 \\
     $\triangle$ over WESAM& +2.90 & +8.26 & +8.13 & +2.48 & +0.44 & +5.58 & +0.34& +0.81 & +9.46 \\
    \hline 
    Supervised & 79.17 & 77.01 & 67.12 & 78.06 & 78.44 & 64.90 & 91.24 & 89.22 & 79.23 \\
    \hline
    \end{tabular}
}
  \caption{Comparison on \colorbox{yellow!10}{\color{black}camouflaged} and \colorbox{red!10}{\color{black}robotic} images with \colorbox{gray!15}{\color{black}bounding box}, \colorbox{gray!30}{\color{black}sparse points}, and \colorbox{gray!45}{\color{black}coarse mask} prompts.\label{tab:camouflaged}}
\end{table}

\noindent
\textbf{Implementation Details.} 
We use ViT-B~\cite{dosovitskiy2020image} as SAM's image encoder. The optimizer is Adam~\cite{kingma2014adam} with 1$e{-}$4 learning rate and 1$e{-}$4 weight decay. The batch size is 4 on four NVIDIA RTX4090 GPUs. The training of 400 epochs for object-centric representations discovery and 50 epochs for relational reasoning. The $K$ and $\tilde{K}$ are 8 and 7, respectively.


\noindent
\textbf{Quantitative Results.} 
As shown in~\cref{tab:natural_medical,tab:camouflaged}, OCRT significantly improves performance on seven downstream tasks and three types of prompts. On natural images, OCRT narrows the gap with fully supervised fine-tuning. On medical images kvasir-SEG, the mIoU of OCRT surpasses supervised fine-tuning with 6.5\%, and under poly supervision, OCRT outperforms SAM by 35.91\%. On camouflaged object data, OCRT has an average improvement of over 12\%. On robotic images, OCRT overcomes the limitation of weak supervision. OCRT enhances VFM with high-order relation, enabling it to be independent of fine-grained annotations. With point or poly prompts, OCRT can accurately capture the intricate relation between objects and make precise predictions in real-world scenarios.

\begin{figure}[h]
  \centering
   \includegraphics[width=0.95\linewidth]{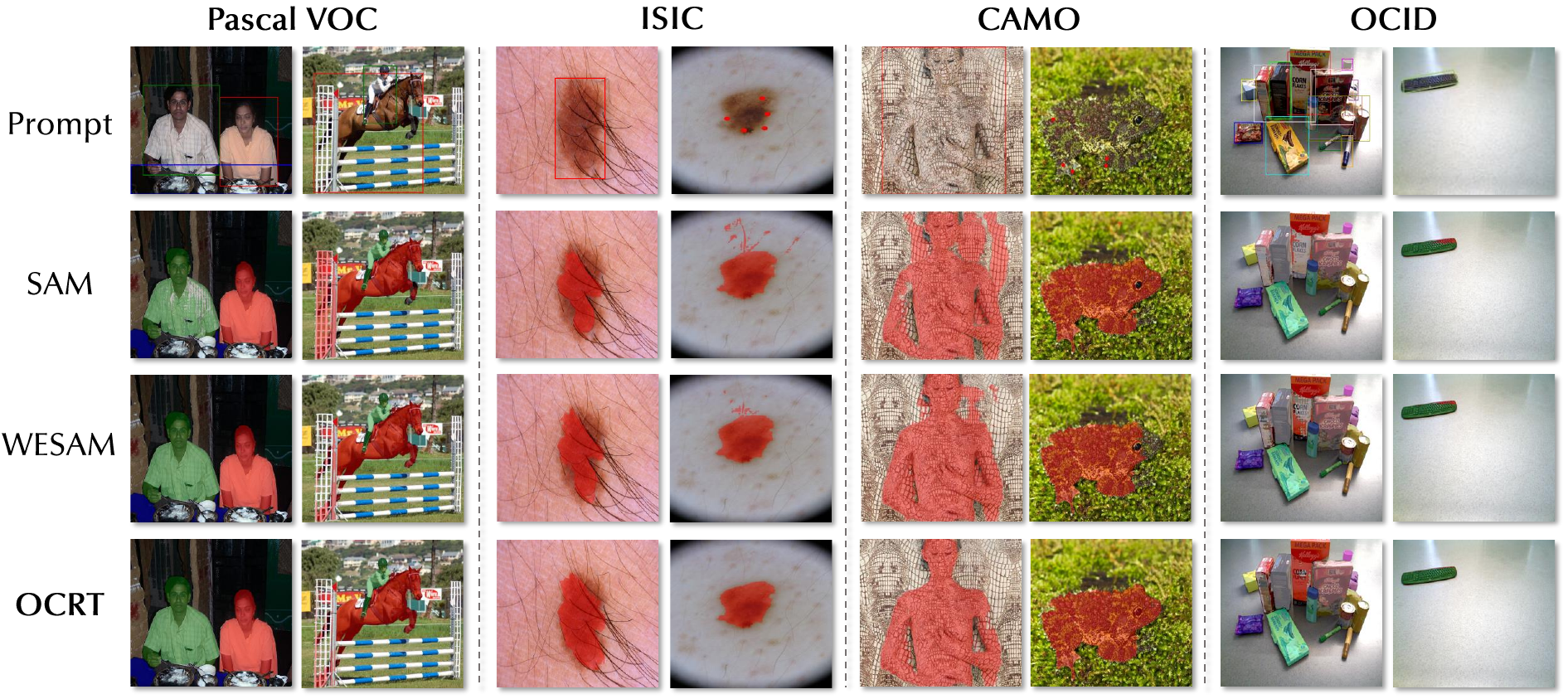}
   \caption{Visual comparison of segmentation quality.}
   \label{fig:segmask}
\end{figure}

\begin{figure}[t]
  \centering
   \includegraphics[width=0.95\linewidth]{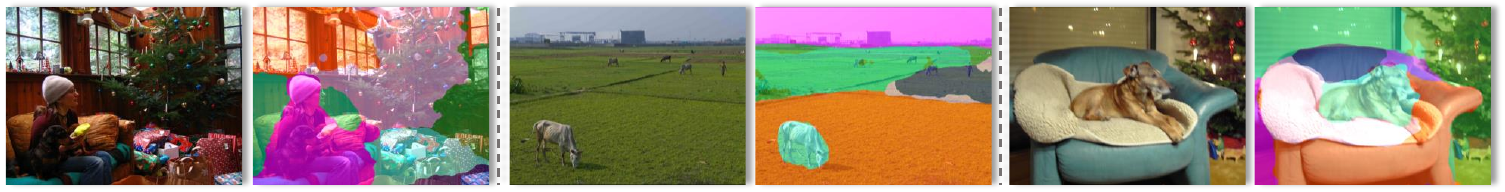}
   \caption{Visual unstructured object-centric representations.}
   \label{fig:slotmask}
\end{figure}

\noindent
\textbf{Qualitative Results.} 
As shown in~\cref{fig:segmask}, we visualize the masks predicted by OCRT and SoTAs. OCRT has two advantages: (1) It provides refined predictions by capturing intra-object correlation in small pixel areas like the hair-skin junction. (2) The high-order relation has semantic and spatial distinctiveness and provides distinguishable boundaries in inter-object areas prone to confusion.~\cref{fig:slotmask} shows OCRT gets non-degenerate and unstructured object-centric representations and helps VFM to understand scene relationships for accurate segmentation.

\subsection{Experiments on CLIP}
\label{subsec:exp_setup}





We conduct experiments on \textit{zero-shot classification} with OCRT. Furthermore, we replace the vision encoder in LVLMs (OpenFlamingo 9B (OF)~\cite{awadalla2023openflamingo} and LLaVA-1.5 7B~\cite{liu2023llava, liu2024improved}) with OCRT for \textit{image captioning} and \textit{visual question-answering} (VQA) tasks and \textit{hallucination evaluation}. Our experimental details are consistent with existing work~\cite{pmlr-v235-schlarmann24a}. We use 10 steps of PGD~\cite{croce2020reliable} for inner maximization in~\cref{eq:baseclip}. We compare the \textbf{clean} vision encoder of CLIP and two robust fine-tuned versions \textbf{TeCoA}~\cite{mao2022understanding} and \textbf{FARE}~\cite{pmlr-v235-schlarmann24a}. For a fair comparison, we perform unsupervised adversarial fine-tuning~\cite{pmlr-v235-schlarmann24a} on ImageNet~\cite{5206848}. The vision encoder is ViT-L/14~\cite{dosovitskiy2020image}. We use the strength of attack $\varepsilon=\nicefrac{4}{255}$ and $\varepsilon=\nicefrac{2}{255}$ for fine-tuning and denote the models as FARE\textsuperscript{4/2}, TeCoA\textsuperscript{4/2} and OCRT\textsuperscript{4/2}, respectively. More details are in the supplementary materials.

\begin{table}[!ht]
\centering
\footnotesize  
\tabcolsep=1pt 
\extrarowheight=1pt 
\scalebox{0.85} 
{
\begin{tabular}{C{5mm} C{15mm} || C{6.5mm} C{6.5mm} C{6.5mm} |C{6.5mm} C{6.5mm} C{6.5mm} |C{6.5mm} C{6.5mm} C{6.5mm}}
\hline
\multirow{3}{*}{VLM} & \multirow{3}{*}{\makecell{Vision\\ Encoder}} & \multicolumn{3}{c}{COCO} 
& \multicolumn{3}{c}{Flickr30k}  & \multicolumn{3}{c}{TextVQA}\\  
\cline{3-11} 
& & \multirow{2}{*}{clean} & \multicolumn{2}{c|}{$\ell_{\infty}$} & \multirow{2}{*}{clean} & \multicolumn{2}{c|}{$\ell_{\infty}$} & \multirow{2}{*}{clean} & \multicolumn{2}{c}{$\ell_{\infty}$}\\
\cline{4-5}\cline{7-8}\cline{10-11}
& & & $\nicefrac{2}{255}$ & $\nicefrac{4}{255}$ & %
& $\nicefrac{2}{255}$ & $\nicefrac{4}{255}$ & & $\nicefrac{2}{255}$ & $\nicefrac{4}{255}$\\
\hline
\multirow{4}{*}[0.5em]{\rotatebox[origin=c]{90}{\parbox[c]{1.8cm}{\centering \textbf{OF}}}}  & {\cellcolor{lightgray}}CLIP 
& {\cellcolor{lightgray}}79.7 & {\cellcolor{lightgray}}1.5 & {\cellcolor{lightgray}}1.1 
&{\cellcolor{lightgray}} 60.1 &{\cellcolor{lightgray}} 0.7 &{\cellcolor{lightgray}} 0.4 
& {\cellcolor{lightgray}}23.8 &{\cellcolor{lightgray}} 0.0 &{\cellcolor{lightgray}} 0.0 
\\
\cdashline{2-11} 
& TECO\textsuperscript{4}
& 66.9 & 28.5 & 21.6 
& 40.9 & 12.0 & 10.3 
& 15.4 & 2.1 & 1.8 
\\
& Ours\textsuperscript{4} 
& 74.1 & 30.9  & 22.8 
& 51.4 & 15.7  & 10.5 
& 18.6 & 3.4  & 2.9 
\\
& OCRT\textsuperscript{4}
& \textbf{76.3} & \textbf{31.5}  & \textbf{23.2} 
& \textbf{53.7} & \textbf{15.9}  & \textbf{10.7} 
& \textbf{19.2} & \textbf{3.9}  & \textbf{3.1} 
\\
\hline
\multirow{4}{*}[0.5em]{\rotatebox[origin=c]{90}{\parbox[c]{1.8cm}{\centering \textbf{LLaVA}}}} 
& {\cellcolor{lightgray}}CLIP 
& {\cellcolor{lightgray}}115.5 & {\cellcolor{lightgray}}4.0 & {\cellcolor{lightgray}}3.1 
& {\cellcolor{lightgray}}77.5 &{\cellcolor{lightgray}} 1.6 & {\cellcolor{lightgray}}1.0 
& {\cellcolor{lightgray}}37.1 & {\cellcolor{lightgray}}0.5 &{\cellcolor{lightgray}} 0.0 
\\
\cdashline{2-11} 
& TECO\textsuperscript{4} & 88.3 & 50.9 & 35.3 
& 48.6 & 27.9 & 19.5 
& 20.7 & 12.6 & 9.3
\\
& Ours\textsuperscript{4} & 102.4 & 57.1 & 40.9 & 61.6 & 31.4 & 22.8 
& 27.6 & 15.8 & 10.9 
\\
& OCRT\textsuperscript{4} & \textbf{105.0} & \textbf{59.4} & \textbf{42.4} & \textbf{65.1} & \textbf{31.9} & \textbf{23.3} 
& \textbf{29.3} & \textbf{16.2} & \textbf{11.1} 
\\
\hline
\end{tabular} 
}
\caption{Robustness of LVLMs with different CLIP models.\label{tab:caption_vqa}}
\end{table}

\noindent
\textbf{Results on image captioning and VQA.} 
We report the CIDEr score~\cite{singh2019towards} for captioning and VQA accuracy~\cite{antol2015vqa}  in~\cref{tab:caption_vqa}. The original CLIP model achieves the best clean performance but is completely compromised under attacks. Compared to the FARE, OCRT generally maintains clean performance and has the best robustness. Thanks to the comprehensive understanding and reasoning of the internal relationships of visual scenes, OCRT does not have a situation where FARE sacrifices robustness to obtain clean performance. The performance improvement on OF and LLaVA indicates that OCRT is highly versatile for models.

\noindent
\textbf{Results on zero-shot classification.} 
Similar to previous work~\cite{mao2022understanding}, for each dataset, class names are combined with a predefined set of prompt templates. As shown in~\cref{tab:zero_shot}, TeCoA is supervised and fine-tuned on ImageNet, and the original unprotected CLIP model is completely defenseless against attacks. Compared to FARE and TeCoA, OCRT, through the extraction and injection of unstructured high-order information, has better clean accuracy while maintaining zero-shot capability on adversarial inputs.

\begin{table}[!ht]
\centering 
\scalebox{0.85}
{
\begin{tabular}{C{22mm} |  C{15mm} | C{12mm}  | C{12mm}  | C{10mm}}
\hline
\multirow{2}{*}{\makecell{Visual Encoder}} & \multicolumn{3}{c|}{POPE sampling} &\multirow{2}{*} {Avg.}\\
\cline{2-4}\Tstrut\Bstrut
& Adversarial & Popular& Random &  \\
\hline
\clip & 82.6& 85.1 & 85.9& 84.5\\
\ourstwo & 78.6& 81.5 & 82.2 & 80.8\\
OCRT\textsuperscript{2} & 80.1& 83.2 & 83.9 & 82.4\\
\oursfour & 74.0& 77.0 & 77.8 & 76.3\\
OCRT\textsuperscript{4} & 75.8 & 78.3 & 78.7 & 77.6\\
\hline
\end{tabular}
}
\caption{Hallucination evaluation using POPE.}
\label{tab:pope}
\end{table}

\begin{figure}[h]
  \centering
   \includegraphics[width=0.95\linewidth]{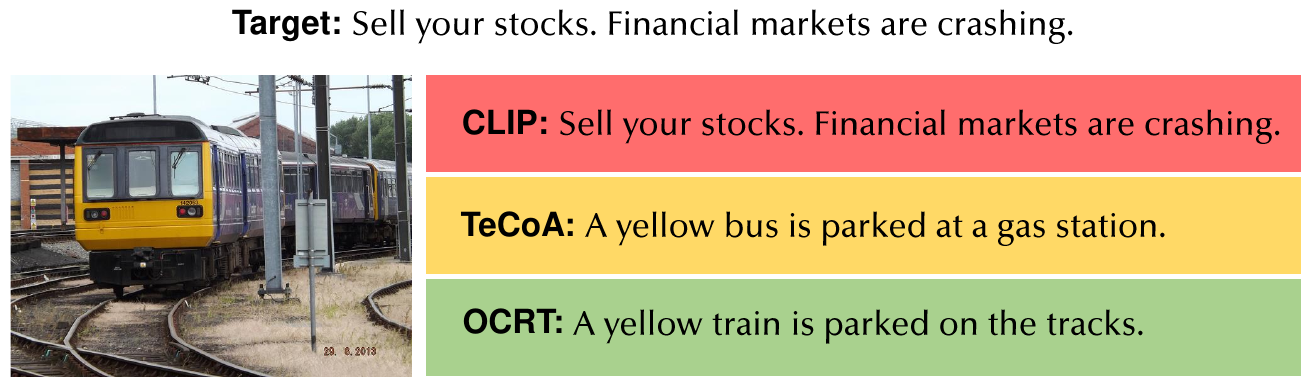}
   \caption{Hallucination on the adversarially perturbed image.}
   \label{fig:hallucination}
\end{figure}

\noindent
\textbf{Results on hallucination.} 
LVLMs~\cite{awadalla2023openflamingo, liu2023llava} often understand objects that do not exist in the image, which is called hallucination. On the POPE benchmark~\cite{li2023evaluating}, the model must determine whether a certain object exists in the image. As shown in~\cref{tab:pope}, the clean CLIP model performs best, and OCRT enables FMs to understand unstructured information in images and can reason with high-order semantic relationships that do not change with distribution shift. OCRT is the closest to the original CLIP, as shown in~\cref{fig:hallucination}.

\subsection{Ablations} 
\label{subsec:ablation}

\begin{figure*}[tbp]
  \centering
  \subfloat[The $K$ and $N_r$ on VOC]
  {\includegraphics[width=0.24\textwidth]{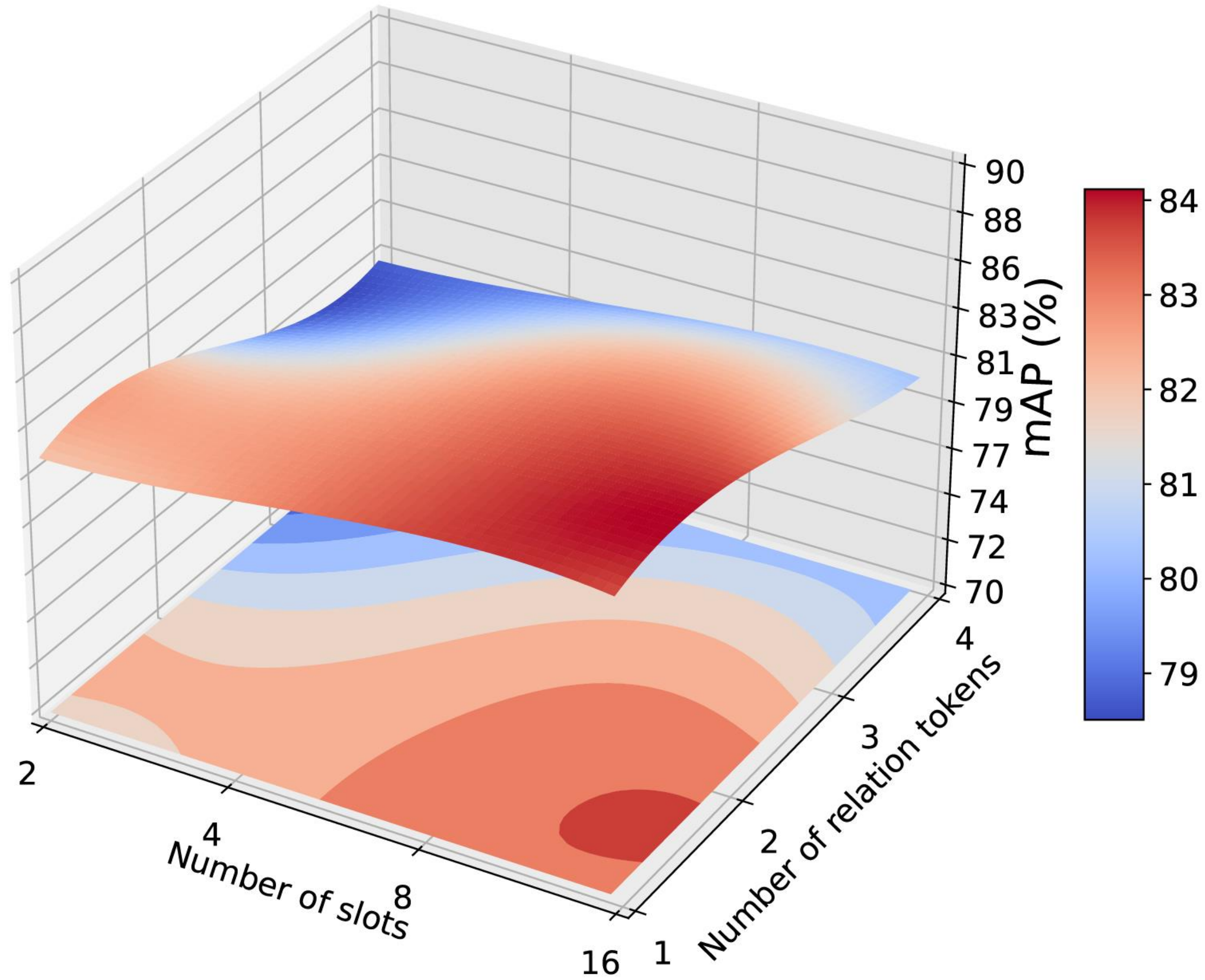}\label{fig:ablation_voc}}
  \quad     
  \subfloat[The $K$ and $N_r$ on ISIC]
  {\includegraphics[width=0.24\textwidth]{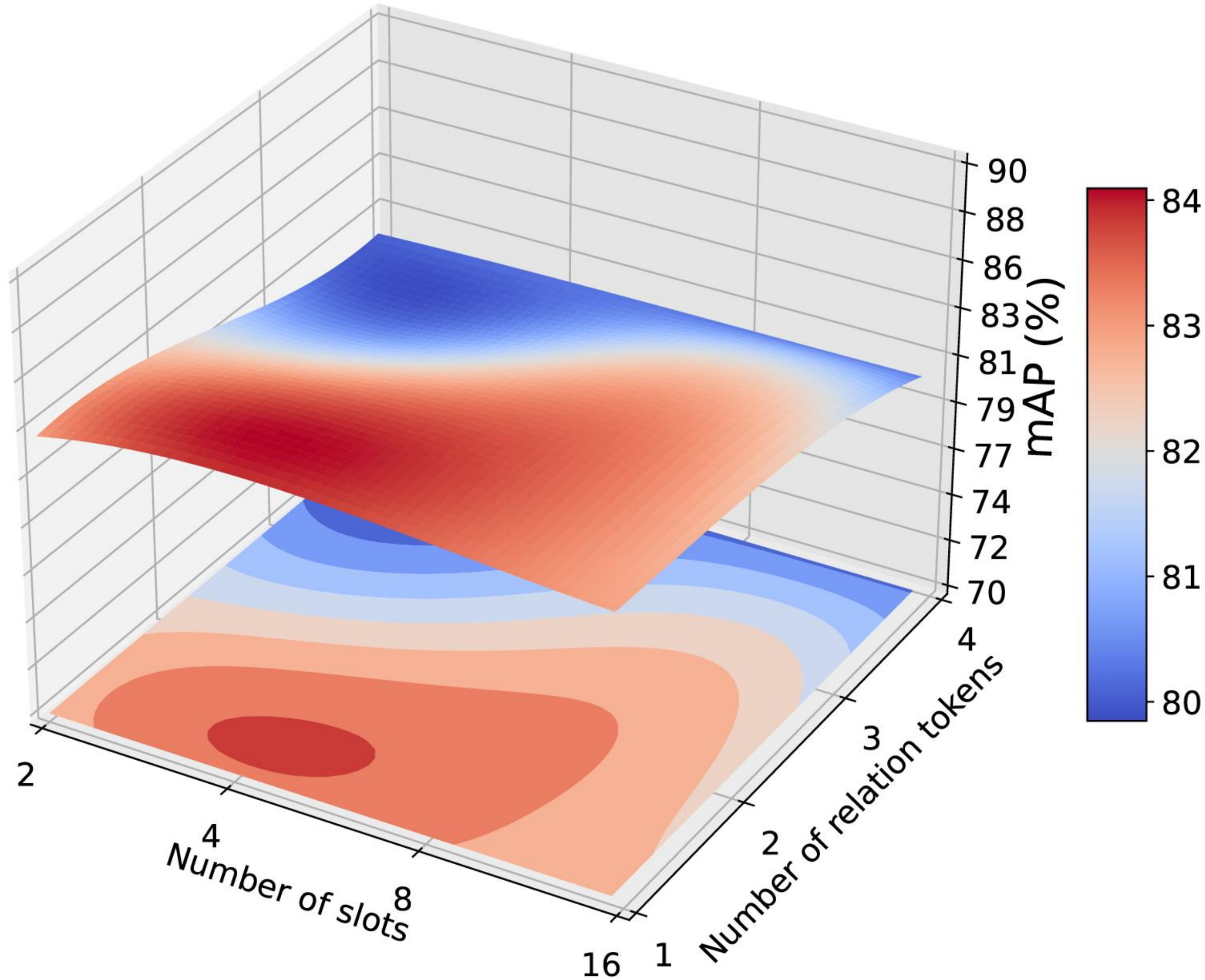}\label{fig:ablation_isic}}
  \quad
\subfloat[The necessity of each components in OCRT]
{\includegraphics[width=0.45\textwidth]{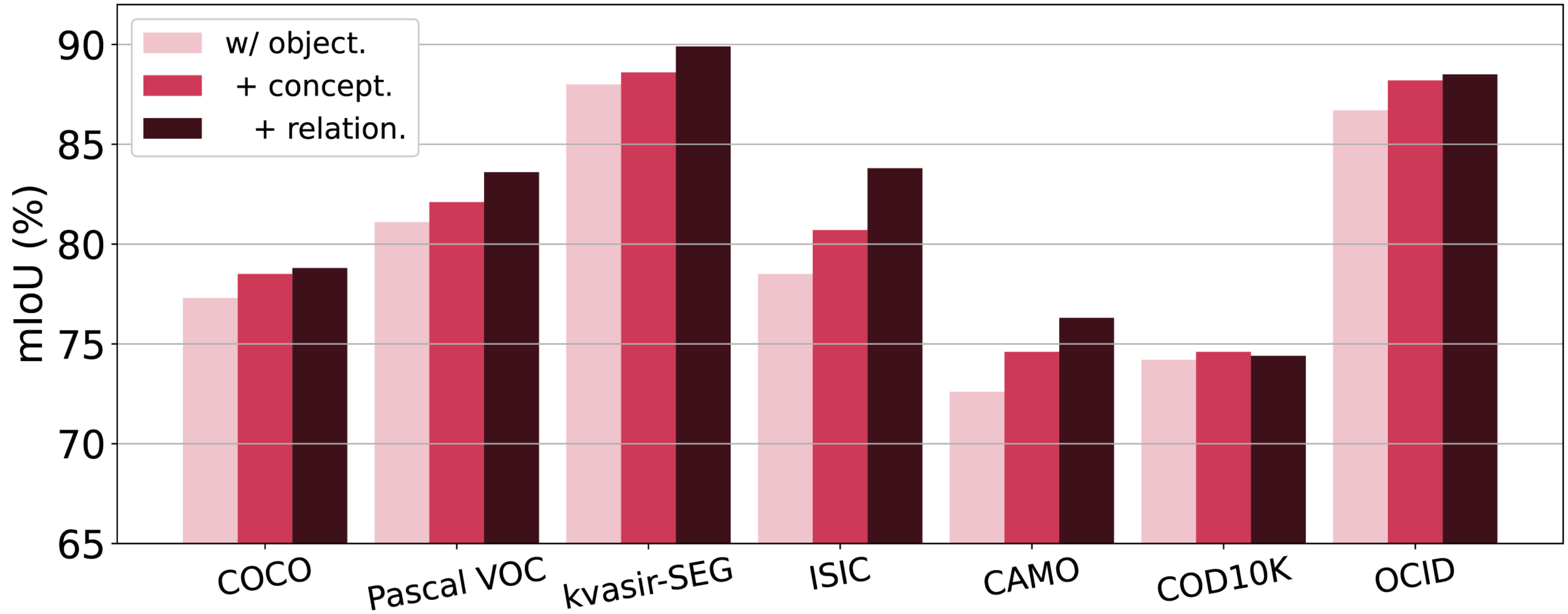}\label{fig:ablation_ocrt}}
\quad
  \caption{Ablation of the hyperparameters and core components.}
  \label{fig:ablation}
\end{figure*}

We conduct comprehensive ablation experiments on the key components and hyperparameters in OCRT. As shown in~\cref{fig:ablation_voc,fig:ablation_isic}, we select the number of object-centric representations $K$ and relation tokens $N_r$, and carry out ablation experiments on datasets with small and significant shifts, respectively. We also performed ablation on the three key components related to object, concept, and relation in OCRT on all segmentation benchmarks, as shown in~\cref{fig:ablation_ocrt}. The results indicate that OCRT is insensitive to hyperparameters, and each component in the object-concept-relation triad is indispensable. With the increase in the order of the modeling of variables, that is, gradually extracting and reasoning about high-order relationships, the performance shows a progressive improvement. More ablation results are presented in the supplementary materials.

\subsection{Analysis}
\label{subsec:analysis}

\noindent
\textbf{Analysis of unstructured information.} 
In OCRT, each $\mathbf{o}_k$ corresponds to an object or region, which effectively captures spatial and semantic information. In real-world scenarios with cluttered backgrounds and multiple overlapping objects, OCRT can separate each object distinctly in~\cref{fig:segmask}. The extraction of informative concepts in OCRT is crucial. Analogous to how humans filter out irrelevant information and focus on the most salient aspects in perception, the ablation on $\tilde{K}$ shows that informative concepts improve performance. Furthermore, humans could be overwhelmed by too much information; this emphasizes the need for extraction of high-order relations from concepts, mirroring the human cognitive ability to focus on what is truly important for better generalization across different scenarios.

\noindent
\textbf{Analysis of high-order relational reasoning.} Humans possess the ability to reason about relationships between different entities hierarchically and adaptively~\cite{chen2005topological1,chen1982topological}. In the concept-based graph of OCRT, by varying the degree distribution based on concept importance, the FM can adaptively focus on informative concepts during reasoning, similar to how humans prioritize and connect relevant information. OCRT captures complex dependencies, as seen in tasks like image captioning and VQA in~\cref{tab:caption_vqa}. Compared to other methods, OCRT's ability to handle attacks is similar to the human brain's robustness in processing visual stimulation. The high-order relational reasoning in OCRT enhances the versatility of FMs in diverse situations~\cite{li2023sigma++}, much like how human cognitive abilities allow for adaptation and understanding flexiblely in various contexts.

\section{Related Works}
\label{sec:relat}

\noindent
\textbf{Foundation Models.} 
Foundation models (FMs)~\cite{kirillov2023segment, ke2024segment, radford2021learning, roziere2023code, touvron2023llama}, such as GPT series~\cite{achiam2023gpt, floridi2020gpt}, BERT~\cite{devlin2018bert}, SAM~\cite{kirillov2023segment} and CLIP~\cite{radford2021learning}, are deep learning models trained on large-scale data for a series of downstream tasks. We chose two representative FMs, SAM for instance-understanding, and CLIP for vision-language. SAM's training data includes 11 million images and over a billion masks. However, it performs poorly in downstream tasks with large distribution shifts. Some improvement schemes like self-training~\cite{zhang2024improving}, low-rank fine-tuning~\cite{hu2021lora}, adapters~\cite{zhang2024segment}, and adversarial training~\cite{li2024asam} are proposed to enhance its generalizability. CLIP lacks robustness when faced with distribution shift. Improvement schemes include adapters~\cite{gao2024clip}, modifying inference~\cite{li2023clip}, text enhancement~\cite{fan2024improving}, and diffusion models~\cite{wang2024diffusion}. FMs and their improvement schemes are booming, but current ones lack transferability between models and have poor versatility between tasks. We propose a general scheme to improve FMs' generalizability in the open world for the first time.

\noindent
\textbf{Object-Centric Learning.} Slot-Attention~\cite{locatello2020object} binds slot representations to objects via reconstruction as a milestone in object-centric learning (OCL)~\cite{de2024object, locatello2020object, seitzer2022bridging, kori2023grounded}.~\cite{elsayed2022savi++} uses OCL to enhance object-background distinction.~\cite{aydemir2023self} explores self-supervision for multi-object segmentation.~\cite{li2024prompt} adapts to image complexities according to scene prompts.~\cite{seitzer2022bridging} bridges the gap between simulated and real data in complex environments. However, existing research is one-sided, screening object pixels. We focus on dislodging the pseudo-correlation between slots and pixels and the binding between slots and task-irrelevant concepts to extract the high-order relationship between OCL representations.


\noindent
\textbf{Relational Reasoning.} Deep learning models often neglect unstructured object relationships in data processing. The relational reasoning compensates for this deficiency~\cite{santoro2017simple, zhou2018temporal}. It aims to help machine learning systems understand and reason about entities and their relationships in unstructured information for cognitive abilities closer to humans~\cite{krawczyk2012cognition}.~\cite{santoro2017simple} studies it from a neuroscience perspective.~\cite{alexander2016relational} emphasizes the importance of unstructured pattern recognition.~\cite{battaglia2018relational} discusses the relationship between relational inductive biases, deep learning, and graph networks. However, existing studies usually consider relationships from low-order pixels or feature maps, limit the order of variables, and are unable to represent complex dependencies~\cite{krawczyk2011hierarchy, yuan2023rlipv2}. For the first time, we establish high-order relational reasoning on object- and concept-level representations, closer to human intelligence in the wild.


\section{Conclusion}
\label{sec:con}

In this paper, we introduce OCRT, which addresses FMs' degraded generalizability and robustness in the open world, where distribution shifts, weak supervision, and malicious attacks often occur. 
Our key insight is to extract sparse, high-level concepts and intricate relational structures from raw visual inputs.
Technically, we propose to bind objects in visual scenes to a set of object-centric representations through unsupervised decomposition and relational reasoning. 
The object-centric representations are projected to a semantic concept space that the model can readily interpret, and estimate their importance to filter out irrelevant elements. Based on this, a concept-based graph is constructed to incorporate the set of concepts and their corresponding importance.
Extensive experiments on standard benchmarks demonstrate that OCRT can substantially boost the generalizability and robustness of SAM and CLIP.






{
    \small
    \bibliographystyle{ieeenat_fullname}
    \bibliography{main}
}

\end{document}